\documentclass[letterpaper, 10 pt, journal, twoside]{IEEEtran}
\let\authorrefmark\IEEEauthorrefmark

\usepackage{style}

\begin{document}
\maketitle
%\thispagestyle{empty}
%\pagestyle{empty}
%\thispagestyle{plain}
%\pagestyle{plain}
%\pagenumbering{roman}

\begin{abstract}
Adaptive control is subject to stability and performance issues when a learned model is used to enhance its performance. This paper thus presents a deep learning-based adaptive control framework for nonlinear systems with multiplicatively-separable parametrization, called adaptive Neural Contraction Metric (aNCM). The aNCM approximates real-time optimization for computing a differential Lyapunov function and a corresponding stabilizing adaptive control law by using a Deep Neural Network (DNN). The use of DNNs permits real-time implementation of the control law and broad applicability to a variety of nonlinear systems with parametric and nonparametric uncertainties. We show using contraction theory that the aNCM ensures exponential boundedness of the distance between the target and controlled trajectories in the presence of parametric uncertainties of the model, learning errors caused by aNCM approximation, and external disturbances. Its superiority to the existing robust and adaptive control methods is demonstrated using a cart-pole balancing model.
\end{abstract}
%\begin{keywords}
%Machine Learning, Robust Adaptive Control, Optimal Control.
%\end{keywords}
\section{Introduction}
\label{introduction}
Future aerospace and robotic exploration missions require that autonomous agents perform complex control tasks in challenging unknown environments while ensuring stability and optimality even for poorly-modeled dynamical systems. Especially when the uncertainties are too large to be treated robustly as external disturbances, real-time implementable adaptive control schemes with provable stability certificates would enhance the autonomous capabilities of these agents.
%%%%%%%%%%%%%%%%%%%%%%%%%%%%%%%%%%%%%%%%%%%%%%%%%%%%%%
%%%%%%%%%%%%%%%%%%%%%%%%%%%%%%%%%%%%%%%%%%%%%%%%%%%%%%
%%%%%%%%%%%%%%%%%%%%%%%%%%%%%%%%%%%%%%%%%%%%%%%%%%%%%%

In this work, we derive a method of adaptive Neural Contraction Metric (aNCM), which establishes a deep learning-based adaptive controller for nonlinear systems with parametric uncertainty. We consider multiplicatively-separable systems in terms of its state $x$ and unknown parameter $\theta$, \ie{}, $f(x,\theta) = Y(x)^{\top}Z(\theta)$, which holds for many types of systems including robotics systems~\cite{Slotine:1228283}, high-fidelity spacecraft dynamics~\cite{doi:10.2514/1.55705}, and systems modeled by basis function approximation or neural networks~\cite{Nelles2001,SannerSlotine1992}. The major advantage of aNCM is its real-time implementability, equipped with contraction-based~\cite{contraction} stability and robustness guarantees even under the presence of such parametric uncertainty, external disturbances, and aNCM learning errors. It also avoids the computation of minimizing geodesics in constructing the adaptive control law, as compared to~\cite{ccm,lopez2019contraction}. Our contributions of presenting the aNCM framework (see Fig.~\ref{ancmdrawing}) are summarized as follows.

This paper builds upon our prior work on Neural Contraction Metrics (NCMs)~\cite{ncm,cdc_ncm,tutorial} for learning-based control and estimation of nonlinear systems. The NCM approximates real-time optimization by utilizing a Deep Neural Network (DNN) to model optimal contraction metrics, the existence of which guarantees exponential boundedness of system trajectories robustly against external disturbances, but without parametric uncertainty. In this study, we newly derive its stability and robustness guarantees explicitly considering the learning error of the NCM, thereby synthesizing a stabilizing real-time adaptive controller for systems with a matched uncertainty condition. Its adaptation law exploits the generalized State-Dependent Coefficient (SDC) parameterization ($A(x,x_d)$ \st{} $A(x,x_d)(x-x_d) = f(x)-f(x_d)$)~\cite{nscm,observer} to provide an exponential bound on the distance between a target trajectory $x_d$ and closed-loop trajectories, while simplifying the differential formulation proposed in~\cite{lopez2019contraction,ccm} that requires the computation of minimizing geodesics. We further generalize this approach to multiplicatively separable systems $f(x,\theta)=Y(x)^{\top}Z(\theta)$ with an unknown constant parameter vector $\theta$, using aNCM to model optimal parameter-dependent contraction metrics along with a novel adaptation law inspired by~\cite{Slotine:1228283} and extending~\cite{lopez2019contraction}. This renders it applicable also to provably stable adaptive control of systems modeled by neural networks and basis function approximation~\cite{Nelles2001,SannerSlotine1992}.

The optimality of aNCM follows from the CV-STEM method~\cite{mypaperTAC} that minimizes a steady-state upper bound of the tracking error perturbed by stochastic and deterministic disturbances by using convex optimization. The NCM method~\cite{ncm,nscm,cdc_ncm,tutorial} samples optimal contraction metrics from CV-STEM to be modeled by a DNN, and is further improved in this paper to incorporate the NCM learning error. In simulation results of the cart-pole balancing task (Fig.~\ref{catpole_fig}), the proposed frameworks are shown to outperform existing adaptive and robust control techniques. Furthermore, the concept of implicit regularization-based adaptation~\cite{boffi2020implicit} can also be incorporated to shape parameter distribution in low excitation or over-parameterized contexts.
%%%%%%%%%%%%%%%%%%%%%%%%%%%%%%%%%%%%%%%%%%%%%%%%%%%%%%
%%%%%%%%%%%%%%%%%%%%%%%%%%%%%%%%%%%%%%%%%%%%%%%%%%%%%%
%%%%%%%%%%%%%%%%%%%%%%%%%%%%%%%%%%%%%%%%%%%%%%%%%%%%%%
\subsubsection*{Related Work}
There exist well-known adaptive stabilization techniques for nonlinear systems equipped with some special structures in their dynamics, \eg{},~\cite{Slotine:1228283,28015,doi:10.1080/00207178608933564,KRSTIC1992177}. They typically construct adaptive control schemes on top of a known Lyapunov function often found based on physical intuition~\cite[p. 392]{Slotine:1228283}. However, finding a Lyapunov function analytically without any prior knowledge of the systems of interest is challenging in general.

Developing numerical schemes for constructing a Lyapunov function has thus been an active field of research~\cite{spencer18lyapunovnn,NIPS2019_8587,8263816,9115021,9146356}. Contraction theory~\cite{contraction} uses a quadratic Lyapunov function of a differential state $\delta x$ (\ie{} $V=\delta x^{\top} M(x)\delta x$) to yield a global and exponential stability result, and convex optimization can be used to construct a contraction metric $M(x)$~\cite{mypaperTAC,AYLWARD20082163,ccm,7989693,quad_stability}. In \cite{lopez2019contraction}, the computed metric is used to estimate unknown system parameters adaptively with rigorous asymptotic stability guarantees, but one drawback is that its problem size grows exponentially with the number of variables and basis functions~\cite{sos_dissertation} while requiring the real-time computation of minimizing geodesics~\cite{ccm}.

We could also utilize over-parameterized mathematical models to approximate the true model and control laws with sampled data~\cite{spencer18lyapunovnn,NIPS2019_8587,ncm,nscm,cdc_ncm,tutorial}. This includes~\cite{8794351}, where a spectrally-normalized DNN is used to model unknown residual dynamics. When the modeling errors are sufficiently small, these techniques yield promising control performance even for general cases with no prior knowledge of the underlying dynamical system. However, poorly-modeled systems with insufficient training data result in conservative stability and robustness certificates~\cite{8794351,boffi2020learning}, unlike the aforementioned adaptive control techniques. Our proposed aNCM integrates the provably stable adaptive control schemes via contraction theory, with the emerging learning-based techniques for real-time applicability~\cite{ncm,nscm,cdc_ncm,tutorial}.

\subsubsection*{Notation}
\label{preliminaries}
For $x \in \mathbb{R}^n$ and $A \in \mathbb{R}^{n \times m}$, we let $\|x\|$, $\delta x$, and $\|A\|$ denote the Euclidean norm, infinitesimal variation of $x$, and induced 2-norm, respectively. We use the notation $A \succ 0$, $A \succeq 0$, $A \prec 0$, and $A \preceq 0$ for positive definite, positive semi-definite, negative definite, and negative semi-definite matrices, respectively, and $\sym(A) = (A+A^{\top})/2$. 
Also, $I_{n} \in \mathbb{R}^{n\times n}$ denotes the identity matrix.
%%%%%%%%%%%%%%%%%%%%%%%%%%%%%%%%%%%%%%%%%
%%%%%%%%%%%%%%%%%%%%%%%%%%%%%%%%%%%%%%%%%
%%%%%%%%%%%%%%%%%%%%%%%%%%%%%%%%%%%%%%%%%
\section{NCM for Trajectory Tracking Control}
\label{sec_ncm}
The Neural Contraction Metric (NCM) is a recently-developed learning-based framework for provably stable and robust feedback control of perturbed nonlinear systems~\cite{ncm,cdc_ncm,tutorial}. In this paper, we explicitly consider the modeling error of the NCM, and present the modified version for tracking control concerning a given target trajectory $(x_d,u_d)$, governed by the following dynamical system with a controller $u \in \mathbb{R}^{m}$:
\begin{align}
\label{sdc_dynamics}
\dot{x} = f(x)+B(x)u+d(x),~\dot{x}_d = f(x_d)+B(x_d)u_d(x_d)
\end{align}
where $x,x_d:\mathbb{R}_{\geq0} \mapsto \mathbb{R}^{n}$, $u_d:\mathbb{R}^n\mapsto\mathbb{R}^{n}$, $d:\mathbb{R}^n\mapsto \mathbb{R}^n$ with $\overline{d}=\sup_{x}\|d(x)\| < +\infty$ is the unknown bounded disturbance, and $f:\mathbb{R}^n \mapsto \mathbb{R}^{n}$ and $B:\mathbb{R}^n\mapsto\mathbb{R}^{n\times m}$ are known smooth functions. Lemma~\ref{sdclemma} is useful for using $(x_d,u_d)$ in the NCM.
\begin{lemma}
\label{sdclemma}
For $f$ and $B$ defined in \eqref{sdc_dynamics}, $\exists A(x,x_d)$ \st{} $f(x)+B(x)u_d(x_d)-f(x_d)-B(x_d)u_d(x_d)=A(x,x_d)(x-x_d),~\forall x,x_d$, and one such $A$ is given as $A(x,x_d) = \int_0^1\bar{f}_x(c x+(1-c)x_d)dc$, where $\bar{f}(q)=f(q)+B(q)u_d(x_d)$ and $\bar{f}_x=\partial \bar{f}/\partial x$. We call $A$ an SDC matrix, and $A$ is non-unique when $n\geq 2$.
%We call $A$ an SDC form when it is constructed to satisfy controllability and observability conditions (see Theorem~\ref{nscm_con} and Corollary~\ref{nscm_est}).
\end{lemma}
\begin{proof}
See~\cite{nscm}.
\end{proof}

We consider the following control law in this section:
\begin{align}
\label{controller}
&u = u_d(x_d)-R(x,x_d)^{-1}B(x)^{\top}\mathcal{M}(x,x_d)(x-x_d)
\end{align}
where $R(x,x_d) \succ 0$ is a weight matrix on the input $u$ and $\mathcal{M}$ is a Deep Neural Network (DNN), called an NCM, learned to satisfy
\begin{align}
\label{Merror}
\|\mathcal{M}(x,x_d)-M(x,x_d)\| \leq \epsilon_{\ell},~\forall x,x_d\in\mathcal{S},~\exists\epsilon_{\ell}\in[0,\infty)
\end{align}
for a compact set $\mathcal{S}\subset\mathbb{R}^n$ and a contraction metric $M$ to be defined in \eqref{convex_opt_ncm}. Let us emphasize that there are two major benefits in using the NCM for robust and adaptive control of nonlinear systems:
\begin{enumerate}
\item Any approximation method could be used to model $M$ as in \eqref{Merror} for its real-time implementability, unlike~\cite{AYLWARD20082163}.
\item $u$ of \eqref{controller} given with $\mathcal{M}$ guarantees stability and robustness even without computing geodesics, unlike~\cite{ccm,lopez2019contraction}.
\end{enumerate}

Theorem~\ref{modified_ncm_thm} presents the modified version of the robust NCM in~\cite{ncm,cdc_ncm,tutorial}, which explicitly considers its modeling error $\epsilon_{\ell}$ and target trajectory $(x_d,u_d)$.
\begin{theorem}
\label{modified_ncm_thm}
Suppose that the contraction metric of \eqref{Merror}, $M(x,x_d)= W(x,x_d)^{-1} \succ 0$  is given by the following convex optimization problem for a given value of $\alpha \in (0,\infty)$:
\begin{align}
    \label{convex_opt_ncm}
    &{J}_{CV}^* = \min_{\nu>0,\chi \in \mathbb{R},\bar{W}\succ 0} ({\overline{d}\chi}/{\alpha_{\rm NCM}}) \text{~~\st{} \eqref{deterministic_contraction_tilde} and \eqref{W_tilde}}
\end{align}
with the convex constraints \eqref{deterministic_contraction_tilde} and \eqref{W_tilde} given as
\begin{align}
\label{deterministic_contraction_tilde}
&-\dot{\bar{W}}+2\sym{\left(A\bar{W}\right)}-2\nu BR^{-1}B^{\top} \preceq -2\alpha \bar{W},~\forall x,x_d \\
\label{W_tilde}
&I_{n} \preceq \bar{W} \preceq \chi I_{n},~\forall x,x_d
\end{align}
where $\underline{\omega},\overline{\omega} \in (0,\infty)$, $\chi = \overline{\omega}/\underline{\omega}$, $\bar{W} = \nu W$, $\nu = 1/\underline{\omega}$, and $\alpha_{\rm NCM}=\alpha - \bar{\rho}\bar{b}^2\epsilon_{\ell}\sqrt{\chi}$. The arguments for $\bar{W}$, $A$, $B$, and $R$ are omitted for notational simplicity, while $B=B(x)$ and $A=A(x,x_d)$ are SDCs of \eqref{sdc_dynamics} given by Lemma~\ref{sdclemma}. Suppose also $\exists \bar{b},\bar{\rho} \in [0,\infty)$ \st{} $\|B(x)\| \leq \bar{b}$ and $\|R^{-1}(x,x_d)\| \leq \bar{\rho},~\forall x,x_d$.

If the NCM modeling error $\epsilon_{\ell}$ of \eqref{Merror} is sufficiently small to satisfy $\alpha_{\rm NCM} > 0$, then the Euclidean distance between $x$ and $x_d$ is exponentially bounded as long as \eqref{sdc_dynamics} is controlled by \eqref{controller}. Furthermore, $M$ minimizes its steady-state upper bound given as ${\overline{d}\chi}/{\alpha_{\rm NCM}}$.
\end{theorem}
\begin{proof}
The virtual system of \eqref{sdc_dynamics} which has $x$ and $x_d$ as its particular solutions is given as $\dot{q}=\dot{x}_d+(A(x,x_d)-B(x)R(x,x_d)^{-1}B(x)^{\top}\mathcal{M}(x,x_d))(q-x_d)+d_q$, where $d_q$ verifies $d_q(x)=d(x)$ and $d_q(x_d)=0$. Thus for a Lyapunov function $V=\int_{x_d}^x\delta q^{\top}M\delta q$, we have using \eqref{Merror} and \eqref{deterministic_contraction_tilde} that
\begin{align}
\dot{V} \leq -2\alpha V+2\int_{x_d}^x\delta q^{\top}M\delta d_q+2\delta q^{\top}MR^{-1}B^{\top}(M-\mathcal{M})\delta q
\end{align}
as in Theorem~2 of~\cite{nscm}. Since the third term is bounded by $2\bar{\rho}\bar{b}^2\epsilon_{\ell}\sqrt{\chi}V$, this gives $\dot{\mathcal{R}} \leq -\alpha_{\rm NCM} \mathcal{R}+\bar{d}/\sqrt{\underline{\omega}}$ for $\mathcal{R}=\int_{x_d}^x\|\Theta\delta q\|$ with $M=\Theta^{\top}\Theta$. The rest follows from the comparison lemma~\cite[pp.102]{Khalil:1173048} as in the proof of Corollary~1 in~\cite{ncm}, as long as $\epsilon_{\ell}$ is small enough to have $\alpha_{\rm NCM} > 0$.
\end{proof}
\begin{figure}
    \centering
    \includegraphics[width=80mm]{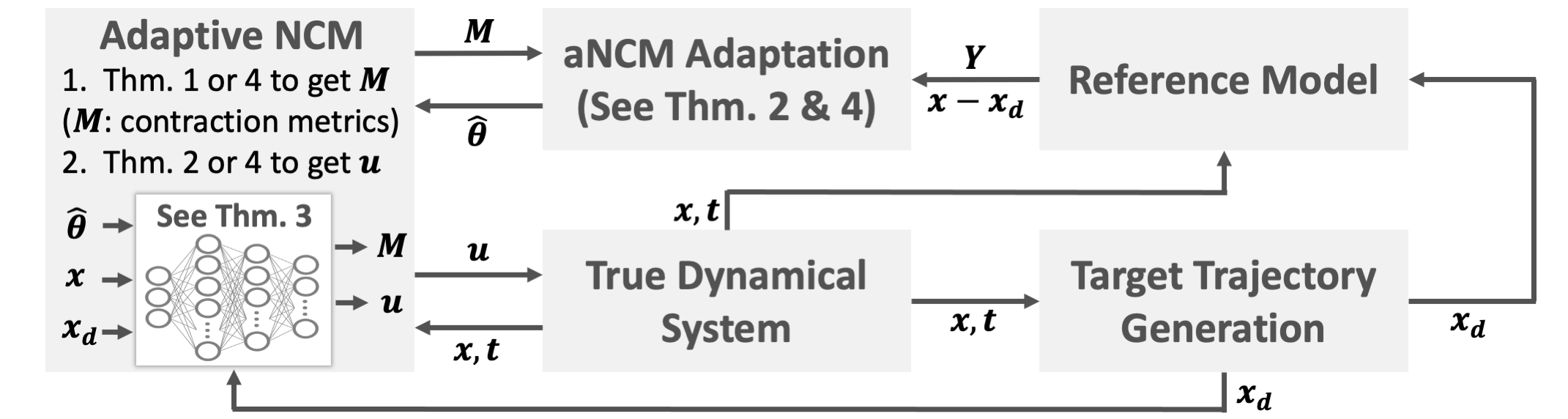}
    \caption{Illustration of aNCM ($M$: aNCM; $\hat{\theta}$: estimated parameter; $Y$: error signal, see \eqref{definition_Y}; $x(t)$ and $x_d(t)$: actual and target state; $u$: control input.}
    \label{ancmdrawing}
    \vspace{-1.4em}
\end{figure}
\section{Adaptive Neural Contraction Metrics}
\label{ancm}
%%%%%%%%%%%%%%%%%%%%%%%%%%%%%%%%%%%%%%%
%%%%%%%%%%%%%%%%%%%%%%%%%%%%%%%%%%%%%%%
%%%%%%%%%%%%%%%%%%%%%%%%%%%%%%%%%%%%%%%
This section elucidates the NCM-based framework for designing real-time adaptive control with formal stability and robustness guarantees of Theorem~\ref{modified_ncm_thm}, as depicted in Fig.~\ref{ancmdrawing}.
%%%%%%%%%%%%%%%%%%%%%%%%%%%%%%%%%%%%%%%
%%%%%%%%%%%%%%%%%%%%%%%%%%%%%%%%%%%%%%%
\subsection{Affine Parametric Uncertainty}
We first consider the following dynamical systems:
\begin{align}
\label{adaptive_affine}
\dot{x} =& f(x)+B(x)u-\Delta(x)^{\top}\theta+d(x) \\
\label{adaptive_affine_d}
\dot{x}_d =& f(x_d)+B(x_d)u_d(x_d)-\Delta(x_d)^{\top}\theta
\end{align}
where $\theta\in\mathbb{R}^p$ is the unknown parameter, $\Delta:\mathbb{R}^n\mapsto\mathbb{R}^{p\times n}$ is a known matrix function, and the other variables are as defined in \eqref{sdc_dynamics}. For these systems with the matched uncertainty condition~\cite{lopez2019contraction}, the NCM in Theorem~\ref{modified_ncm_thm} can be utilized to design its adaptive counterpart.
\begin{theorem}
\label{adaptive_affine_thm}
Suppose $M(x,x_d)$ of \eqref{convex_opt_ncm} is constructed with an additional convex constraint $\partial_{b_i(x)}\bar{W}+\partial_{b_i(x_d)}\bar{W}=0$, where $\partial_{v(q)}\bar{W}=\sum_{i}(\partial \bar{W}/\partial q_i)v_i$ and $B(x)=[b_1(x),\cdots,b_m(x)]$~\cite{ccm,lopez2019contraction}, for the nominal system (\ie{} \eqref{adaptive_affine} and \eqref{adaptive_affine_d} with $\Delta(x,x_d)=0,\forall x,x_d$), and let $\mathcal{M}(x,x_d)$ be an NCM of \eqref{Merror} in Theorem~\ref{modified_ncm_thm} with such $M$. Suppose also that the matched uncertainty condition~\cite{lopez2019contraction} holds, \ie{} $(\Delta(x)-\Delta(x_d))^{\top}\theta \in \text{span}(B(x))$, and that \eqref{adaptive_affine} is controlled by the following adaptive control law:
\begin{align}
\label{affine_adaptive_u}
u =& u_d(x_d)-R(x,x_d)^{-1}B(x)^{\top}\mathcal{M}(x-x_d)+\varphi(x,x_d)^{\top} \hat{\theta} \\
\label{affine_adaptation}
\dot{\hat{\theta}} =& -\Gamma (\varphi(x,x_d)B(x)^{\top}\mathcal{M}(x-x_d)+\sigma \hat{\theta})
\end{align}
where $(\Delta(x)-\Delta(x_d))^{\top}\theta=B(x)\varphi(x,x_d)^{\top}\theta$, $\Gamma \succ 0$, $\sigma\in[0,\infty)$, and the arguments of $\mathcal{M}$ are omitted for notational simplicity. If $\exists \underline{\gamma},\overline{\gamma},\bar{b},\bar{\rho},\bar{\phi},\bar{\theta} \in (0,\infty)$ \st{} $\underline{\gamma} I_p \preceq \Gamma \preceq \overline{\gamma}I_p$, $\|B(x,\theta)\| \leq \bar{b}$, $\|R^{-1}(x,x_d)\| \leq \bar{\rho}$, $\|\varphi(x,x_d)\| \leq \bar{\phi},\forall x,x_d,\theta$, and $\|\theta\| \leq \bar{\theta}$, and if $\Gamma$ and $\sigma$ of \eqref{affine_adaptation} are selected to satisfy the following relation for the learning error $\epsilon_{\ell}$ of \eqref{Merror}:
\begin{align}
\label{condition_mNCM}
\begin{bmatrix}
-2\alpha_{\mathrm{NCM}}/\overline{\omega} & \bar{\phi}\bar{b}\epsilon_{\ell} \\
\bar{\phi}\bar{b}\epsilon_{\ell} & -2\sigma
\end{bmatrix} \preceq -2 \alpha_{a}\begin{bmatrix}
1/\underline{\omega} & 0 \\
0 & 1/\underline{\gamma}
\end{bmatrix}
\end{align}
for $\exists \alpha_{a} \in (0,\infty)$, where $\alpha_{\mathrm{NCM}}$, $\underline{\omega}$, and $\overline{\omega}$ are given in Theorem~\ref{modified_ncm_thm}, we have the following bound:
\begin{align}
\label{adaptive_bound_1}
\|\mathtt{e}(t)\| \leq \sqrt{\overline{\omega}}(\sqrt{V(0)}e^{-\alpha_a t}+\alpha_a^{-1}\bar{d}_{a}(1-e^{-\alpha_a t}))
\end{align}
where $\mathtt{e}=x-x_d$, $\tilde{\theta}=\hat{\theta}-\theta$, $V(t) = \mathtt{e}^{\top}M(x,x_d)\mathtt{e}+\tilde{\theta}^{\top}\Gamma^{-1}\tilde{\theta}$, and $\bar{d}_{a} = \sigma\sqrt{\overline{\gamma}}\bar{\theta}+\bar{d}/\sqrt{\underline{\omega}}$ for $\bar{d} = \sup_{x}\|d(x)\|$ in \eqref{adaptive_affine}.
\end{theorem}
\begin{proof}
Let $A_{cl}=A(x,x_d)-B(x)R(x,x_d)^{-1}B(x)^{\top}\mathcal{M}$. Since the dynamics of $\mathtt{e}$ with $u$ of \eqref{affine_adaptive_u} is given as $\dot{\mathtt{e}}=A_{cl}\mathtt{e}+B(x)\varphi(x,x_d)^{\top}\tilde{\theta}+d(x)$ by the relation $(\Delta(x)-\Delta(x_d))^{\top}\theta=B(x)\varphi(x,x_d)^{\top}\theta$, the condition $\partial_{b_i(x)}\bar{W}+\partial_{b_i(x_d)}\bar{W}=0$, or equivalently, $\partial_{b_i(x)}M+\partial_{b_i(x_d)}M=0$~\cite{ccm} yields
\begin{align}
\dot{V}/2 \leq -\alpha_{\rm NCM}\mathtt{e}^{\top}M\mathtt{e}+\mathtt{e}^{\top}(M-\mathcal{M})B\varphi^{\top}\tilde{\theta}-\sigma\tilde{\theta}^{\top}\hat{\theta}+\mathtt{e}^{\top}Md \nonumber
\end{align}
for $V$ in \eqref{adaptive_bound_1} as in the proof of Theorem~\ref{modified_ncm_thm}, where the adaptation law \eqref{affine_adaptation} is used for $\dot{\hat{\theta}} = \dot{\tilde{\theta}}$. Applying \eqref{Merror} and \eqref{condition_mNCM} with the inequalities $\overline{\omega}^{-1}I_n \preceq M \preceq \underline{\omega}^{-1}I_n$ and $-\sigma\tilde{\theta}^{\top}\hat{\theta}+\mathtt{e}^{\top}Md \leq -\sigma\|\tilde{\theta}\|^2+\bar{d}_{a}\sqrt{V}$ for $\bar{d}_{a}$ defined in \eqref{adaptive_bound_1}, we get
\begin{align}
\dot{V}/2 \leq& -(\alpha_{\mathrm{NCM}}/\overline{\omega})\|\mathtt{e}\|^2+\bar{\phi}\bar{b}\epsilon_{\ell}\|\mathtt{e}\|\|\tilde{\theta}\|-\sigma\|\tilde{\theta}\|^2+\bar{d}_{a}\sqrt{V} \nonumber \\
\leq& -\alpha_{a}(\|\mathtt{e}\|^2/\underline{\omega}+\|\tilde{\theta}\|^2/\underline{\gamma})+\bar{d}_{a}\sqrt{V} \leq -\alpha_{a}V+\bar{d}_{a}\sqrt{V}
\end{align}
which results in $d\sqrt{V}/dt \leq -\alpha_{a}\sqrt{V}+\bar{d}_{a}$. The comparison lemma~\cite[pp.102]{Khalil:1173048} with $\|\mathtt{e}\| \leq \sqrt{\overline{\omega}}\sqrt{V}$ gives \eqref{adaptive_bound_1}.
\end{proof}

Asymptotic stability using Barbalat's lemma as in standard adaptive control is also obtainable when $\epsilon_{\ell} = 0$.
\begin{corollary}
\label{Cor:adaptiveNCM}
The NCM adaptive control \eqref{affine_adaptive_u} with the adaptation \eqref{affine_adaptation} guarantees $\lim_{t\to\infty}\|\mathtt{e}(t)\| = 0$ for $\mathtt{e}=x-x_d$ when $\epsilon_{\ell} = 0$, $d(x) = 0$, and $\sigma = 0$ in \eqref{Merror}, \eqref{adaptive_affine}, and \eqref{affine_adaptation}.
\end{corollary}
\begin{proof}
For $V$ in \eqref{adaptive_bound_1}, we have $\dot{V}/2 \leq -\alpha\mathtt{e}^{\top}M\mathtt{e}+\mathtt{e}^{\top}MB\varphi^{\top}\tilde{\theta}+\tilde{\theta}^{\top}\Gamma^{-1}\dot{\tilde{\theta}} = -\alpha\mathtt{e}^{\top}M\mathtt{e}$ by \eqref{affine_adaptation} with $\sigma=0$. The application of Barbalat's lemma~\cite[pp. 323]{Khalil:1173048} as in the proof of Theorem~2 in~\cite{lopez2019contraction} gives $\lim_{t\to\infty}\|\mathtt{e}(t)\| = 0$.
\end{proof}
\begin{remark}
\label{Remark:objective}
The steady-state error of \eqref{adaptive_bound_1} could be used as the objective function of \eqref{convex_opt_ncm}, regarding $\Gamma$ and $\sigma$ as decision variables, to get $M$ optimal in a sense different from Theorem~\ref{modified_ncm_thm}. Smaller $\epsilon_{\ell}$ would lead to a weaker condition on them in \eqref{condition_mNCM}. Also, the size of $\|\theta\| \leq \bar{\theta}$ in \eqref{adaptive_bound_1} can be adjusted simply by rescaling it (\eg{}, $\theta\to\theta/\bar{\theta}$).
\end{remark}
%%%%%%%%%%%%%%%%%%%%%%%%%%%%%%%%%%%%%%%
%%%%%%%%%%%%%%%%%%%%%%%%%%%%%%%%%%%%%%%
\subsection{NCM for Lagrangian-type Nonlinear Systems}
We have thus far examined the case where $f(x)$ is affine in its parameter. This section considers the following dynamical system with an uncertain parameter $\theta$ and a control input $\tau$:
\begin{align}
\label{lagrangian_dynamics}
H(s)\dot{s}+h(s)+\Delta(s)\theta = \tau +d(s)
\end{align}
where $s\in \mathbb{R}^n$, $\tau\in \mathbb{R}^n$, $H:\mathbb{R}^{n}\mapsto \mathbb{R}^{n\times n}$, $h:\mathbb{R}^{n}\mapsto \mathbb{R}^{n}$, $\Delta:\mathbb{R}^{n}\mapsto \mathbb{R}^{n\times p}$, $d:\mathbb{R}^{n}\mapsto \mathbb{R}^{n}$ with $\bar{d}_s=\sup_s\|d(s)\| < \infty$, and $H(s)$ is non-singular for all $s$. We often encounter the problem of designing $\tau$ guaranteeing exponential boundedness of $s$, one example of which is the tracking control of Lagrangian systems~\cite{Slotine:1228283}. The NCM is also applicable to such problems.
\begin{theorem}
\label{lagrangian_NCM}
Let $\mathcal{M}(s)$ be an NCM for the system $\dot{s}=-H(s)^{-1}h(s)+H(s)^{-1}\tau+H(s)^{-1}d$ given by Theorem~\ref{modified_ncm_thm} with an additional convex constraint $\partial_{b_i(s)}\bar{W}=0$~\cite{ccm,lopez2019contraction} for $B(s)=H(s)^{-1}=[b_1(s),\cdots,b_n(s)]$. Suppose $\tau$ is designed as
\begin{align}
\label{lagrange_control}
\tau = -R^{-1}H^{-\top}\mathcal{M}s+\Delta \hat{\theta},~\dot{\hat{\theta}} = -\Gamma(\Delta ^{\top}H^{-\top}\mathcal{M}s+\sigma s)
\end{align}
where $\Gamma \succ 0$, $\sigma \in [0,\infty)$, $R(s)\succ0$ is a given weight matrix on $\tau$, and the arguments are suppressed for notational convenience. If $\exists \bar{b},\bar{\rho},\bar{\delta} \in (0,\infty)$ \st{} $\|B(s)\| \leq \bar{b}$, $\|R^{-1}(s)\| \leq \bar{\rho}$, and $\|\Delta(s)\| \leq \bar{\delta},\forall s$, and if $\sigma$ and $\Gamma$ of \eqref{affine_adaptation} are selected to satisfy \eqref{condition_mNCM} with $\bar{\phi}=\bar{\delta}$, then we have the exponential bound \eqref{adaptive_bound_1} with $\mathtt{e} = s$, $\bar{d}=\bar{b}\bar{d}_s$, and $V = s^{\top}M(s)s+\tilde{\theta}^{\top}\Gamma^{-1}\tilde{\theta}$.
\end{theorem}
\begin{proof}
Using $\partial_{b_i(s)}\bar{W}=0$ and \eqref{lagrange_control}, we get $\dot{V}/2\leq-\alpha_{\mathrm{NCM}} s^{\top}Ms+s^{\top}(M-\mathcal{M})H^{-1}\Delta\tilde{\theta}-\sigma\tilde{\theta}^{\top}\hat{\theta}+s^{\top}MH^{-1}d$ as in Theorem~\ref{adaptive_affine_thm}. Thus, we have $d\sqrt{V}/dt \leq -\alpha_a\sqrt{V}+\bar{d}_a$ for $\bar{d}_a = \sigma\sqrt{\overline{\gamma}}\bar{\theta}+\bar{b}\bar{d}_s/\sqrt{\underline{\omega}}$ due to \eqref{condition_mNCM}, resulting in \eqref{adaptive_bound_1}.
\end{proof}
\begin{remark}
When $\epsilon_{\ell} = 0$, $d(x) = 0$, and $\sigma = 0$, \eqref{lagrange_control} reduces to asymptotic stabilization of \eqref{lagrangian_dynamics} as in Corollary~\ref{Cor:adaptiveNCM}.
\end{remark}
%%%%%%%%%%%%%%%%%%%%%%%%%%%%%%%%%%%%%%%
%%%%%%%%%%%%%%%%%%%%%%%%%%%%%%%%%%%%%%%
\subsection{Multiplicatively-Separable Parametric Uncertainty}
Next, let us consider the following nonlinear system with an uncertain parameter $\theta\in\mathbb{R}^p$ in \eqref{sdc_dynamics}:
\begin{align}
\label{adaptive_general}
\dot{x} =& f(x,\theta)+B(x,\theta)u+d(x) \\
\label{adaptive_general_d}
\dot{x}_d =& f(x_d,\theta)+B(x_d,\theta)u_d(x_d).
\end{align}
In this section, we assume the following.
\begin{assumption}
\label{multiplicative_asmp}
The dynamical systems \eqref{adaptive_general} and \eqref{adaptive_general_d} are multiplicatively-separable in terms of $x$ and $\theta$, \ie{}, $\exists$ $Y_f:\mathbb{R}^n\mapsto\mathbb{R}^{n\times q_z}$, $Y_{b_i}:\mathbb{R}^n\mapsto\mathbb{R}^{n\times q_z},\forall i$, and $Z:\mathbb{R}^p\mapsto\mathbb{R}^{q_z}$ \st{} 
\begin{align}
\label{linear_adaptive}
Y_f(x)Z(\theta) = f(x,\theta),~Y_{b_i}(x)Z(\theta) = b_i(x,\theta),~\forall x,\theta
\end{align}
where $B(x,\theta)=[b_1(x,\theta),\cdots,b_m(x,\theta)]$. 
\end{assumption}
\begin{remark}
\label{bregman:remark}
When \eqref{linear_adaptive} holds, we could redefine $\theta$ as $[\theta^{\top},Z(\theta)^{\top}]^{\top}$ to get $Y_f(q)\theta = f(q,\theta)$ and $Y_{b_i}(q)\theta = b_i(q,\theta)$. Since such $\theta$ can be regularized as in~\cite{boffi2020implicit} (see Sec.~\ref{Sec:bregman}), we denote $[\theta^{\top},Z(\theta)^{\top}]^{\top}$ as $\theta$ in the subsequent discussion.
\end{remark}

Under Assumption~\ref{multiplicative_asmp} with $\theta$ augmented as $[\theta^{\top},Z(\theta)^{\top}]^{\top}$, the dynamics for $\mathtt{e}=x-x_d$ is expressed as follows: 
\begin{align}
\label{error_dynamics}
\dot{\mathtt{e}} =& A(x,x_d;\hat{\theta})\mathtt{e}+B(x;\hat{\theta})(u-u_d(x_d))-\tilde{Y}(\hat{\theta}-\theta)+d(x) \\
\label{definition_Y}
\tilde{Y} =& Y-Y_d= (Y_f(x)+Y_b(x,u))-(Y_f(x_d)+Y_b(x_d,u_d))
\end{align}
where $Y_b(x,u)=\sum_{i=1}^mY_{b_i}(q)u_i$, $u_d = u_d(x_d)$, $A$ is the SDC matrix in Lemma~\ref{sdclemma}, and $\hat{\theta}$ is the estimate of $\theta$. We design the adaptive control law for \eqref{adaptive_general} as follows:
\begin{align}
\label{adaptive_general_u}
u =& u_d(x_d)-R(x,x_d)^{-1}B(x,\hat{\theta})^{\top}\mathcal{M}(x-x_d)\\
\label{adaptation_general}
\dot{\hat{\theta}} =& \Gamma((Y^{\top}d\mathcal{M}_x+Y_d^{\top}d\mathcal{M}_{x_d}+\tilde{Y}^{\top}\mathcal{M}) (x-x_d)-\sigma \hat{\theta})
\end{align}
where $d\mathcal{M}_q = [({\partial \mathcal{M}}/{\partial q_1})\mathtt{e},\cdots,({\partial \mathcal{M}}/{\partial q_n})\mathtt{e}]^{\top}/2$, $\Gamma \succ 0$, $\sigma \in [0,\infty)$, $Y$, $Y_d$, and $\tilde{Y}$ are given in \eqref{definition_Y}, $R(x,x_d)\succ0$ is a weight matrix on $u$, and $\mathcal{M}=\mathcal{M}(x,x_d,\hat{\theta})$ is a DNN, called an adaptive NCM (aNCM), learned to satisfy
\begin{align}
\label{Merror_adaptive_d}
&\|d\mathcal{M}_q(x,x_d,\hat{\theta})-dM_q(x,x_d,\hat{\theta})\| \leq \epsilon_{\ell} \\
\label{Merror_adaptive}
&\|\mathcal{M}(x,x_d,\hat{\theta})-M(x,x_d,\hat{\theta})\| \leq \epsilon_{\ell},~\forall x,x_d\in\mathcal{S},~\hat{\theta}\in\mathcal{S}_{\theta}
\end{align} 
both for $q=x$ and $x_d$, where $\mathcal{S}\subset\mathbb{R}^n$ and $\mathcal{S}\subset\mathbb{R}^n$ are some compact sets and $M$ is a contraction metric $M$ to be defined in \eqref{convex_opt_ancm}. Theorem~\ref{adaptiveNCMthm} derives a stability guarantee of \eqref{adaptive_general_u}.
\if0
The following lemma is useful for stability analysis of \eqref{error_dynamics}.
\begin{lemma}
\label{lyapunov_eMe}
For a Lyapunov function $V(x,x_d)=R(x,x_d)^2$ where $R$ is continuously differentiable, $\exists M(x,x_d)$ \st{} $V(x,x_d)=e^{\top}M(x,x_d)e$.
\end{lemma}
\begin{proof}
As in Lemma~\ref{sdclemma}, $\exists \mu$ \st{} $R=\mu (x-x_d)=(\int_0^1R_x(cx+(1-c)x_d,x_d)dc)(x-x_d)$. Defining $M$ as $M = \mu^{\top}\mu$ gives the desired relation.
\end{proof}
Lemma~\ref{lyapunov_eMe} indicates that the problem of finding a Lyapunov function $V(x,x_d)$ reduces to the problem of finding a metric $M(x,x_d)$,
\fi
%\begin{remark}
%Note that the conditions \eqref{linear_adaptive} and \eqref{linear_adaptive2} are not strict in a sense that we can always over-parameterize the system, and the adaptation law with the Bregman divergence~\cite{boffi2020implicit} will give us some insight on the implicit regularization on these parameters, which could circumvent the issue of overfitting.
%\end{remark}
%\begin{example}
%For $f(x,\theta) = x_1\theta+x_3^2\theta^2+x_2^3\cos\theta$, we have
%\begin{align}
%f(x,\theta) = [x_1~x_3^2~x_2^3]\begin{bmatrix}\theta_{\max} & 0 & 0\\0 & \theta_{\max}^2 & 0 \\ 0 & 0 & 1\end{bmatrix}\begin{bmatrix}\theta/\theta_{\max}\\\theta^2/\theta_{\max}^2 \\ \cos\theta\end{bmatrix}
%\end{align}
%so we can use $\bar{\theta} = [\theta/\theta_{\max}~\theta^2/\theta_{\max}^2~\cos\theta]^{\top}$ as the unknown parameter.
%\end{example}
\begin{theorem}
\label{adaptiveNCMthm}
Suppose that Assumption~\ref{multiplicative_asmp} holds and let $B=B(x;\hat{\theta})$ and $A=A(x,x_d;\hat{\theta})$ in \eqref{error_dynamics} for notational simplicity. Suppose also $M(x,x_d,\hat{\theta}) = W(x,x_d,\hat{\theta})^{-1} \succ 0$ of \eqref{Merror_adaptive} is given by the following convex optimization for given $\alpha \in (0,\infty)$:
\begin{align}
    \label{convex_opt_ancm}
    &{J}_{aCV}^* = \min_{\nu>0,\chi \in \mathbb{R},\bar{W}\succ 0} ({\overline{d}\chi}/{\alpha_{\mathrm{NCM}}}) \text{~~\st{} \eqref{deterministic_contraction_adaptive_tilde} and \eqref{adaptive_W_tilde}}.
\end{align}
with the convex constraints \eqref{deterministic_contraction_adaptive_tilde} and \eqref{adaptive_W_tilde} given as
\begin{align}
\label{deterministic_contraction_adaptive_tilde}
&-\left.({d}/{dt})\right|_{\hat{\theta}}{\bar{W}}+2\sym{\left(A\bar{W}\right)}-2\nu BR^{-1}B^{\top} \preceq -2\alpha \bar{W} \\
\label{adaptive_W_tilde}
&I_{n} \preceq \bar{W} \preceq \chi I_{n},~\forall x,x_d,\hat{\theta}
\end{align}
where $\underline{\omega}$, $\overline{\omega}$, $\chi$, $\bar{W}$, and $\nu$ are given in \eqref{convex_opt_ncm}, $({d}/{dt})|_{\hat{\theta}}{\bar{W}}$ is the time derivative of $\bar{W}$ computed along \eqref{adaptive_general} and \eqref{adaptive_general_d} with $\theta = \hat{\theta}$, and $\alpha_{\rm NCM}=\alpha - \bar{\rho}\bar{b}^2\epsilon_{\ell}\sqrt{\chi}$ is constructed with $\epsilon_{\ell}$ of \eqref{Merror_adaptive} and \eqref{Merror_adaptive_d} to satisfy $\alpha_{\rm NCM}>0$. Note that the arguments for $\bar{W}$ and $R$ are also omitted for simplicity. If $\exists \bar{b},\bar{\rho},\bar{y} \in (0,\infty)$ \st{} $\|B(x,\theta)\| \leq \bar{b}$, $\|R^{-1}(x,x_d)\| \leq \bar{\rho}$, $\|Y\|\leq \bar{y}$, $\|Y_d\|\leq \bar{y}$, and $\|\tilde{Y}\|\leq \bar{y},~\forall x,x_d,\theta$ in \eqref{adaptive_general_u} and \eqref{adaptation_general}, and if $\Gamma$ and $\sigma$ of \eqref{adaptation_general} are selected to satisfy the following for $\epsilon_{\ell}$ of \eqref{Merror_adaptive} and \eqref{Merror_adaptive_d}:
\begin{align}
\begin{bmatrix}
\label{condition_aNCM}
-2\alpha_{\mathrm{NCM}}/\overline{\omega} & \bar{y}\epsilon_{\ell} \\
\bar{y}\epsilon_{\ell} & -2\sigma
\end{bmatrix} \preceq -2 \alpha_{a}\begin{bmatrix}
1/\underline{\omega} & 0 \\
0 & 1/\underline{\gamma}
\end{bmatrix}
\end{align}
for $\exists \alpha_a\in(0,\infty)$, then we have the exponential bound \eqref{adaptive_bound_1} as long as \eqref{adaptive_general} is controlled by the aNCM control of \eqref{adaptive_general_u}.
\end{theorem}
\begin{proof}
%The virtual system of \eqref{adaptive_general} and \eqref{adaptation_general} with \eqref{adaptive_general_u}, which has $(x,\theta)$ and $(x_d,\hat{\theta})$ as its particular solutions is given as
%\begin{align}
%\label{virtual_q1_general}
%\dot{q}_1 =& \dot{x}_d+A_{cl}(q_1-x_d)-\tilde{Y}(q_2-\theta) \\
%\label{virtual_q2_general}
%\dot{q}_2 =& \Gamma^{-1}(Y^{\top}dM_x(e)+Y_d^{\top}dM_{x_d}(e)+\tilde{Y}^{\top}M) (q_1-x_d)
%\end{align}
%where $A_{cl}=A(x,x_d,t;\hat{\theta})-B(x,t;\hat{\theta})B(x,t;\hat{\theta})^{\top}M(x,x_d,\hat{\theta},t)$.
Since we have $\sum_{i=1}^n({\partial M}/{\partial q_i})\dot{q}_i\mathtt{e}=2d\mathcal{M}_{q}\dot{q}$ for $q=x$ and $q=x_d$, computing $\dot{M}\mathtt{e}$ along \eqref{adaptive_general} and \eqref{adaptive_general_d} yields
\begin{align}
\dot{M}\mathtt{e}=&\left(\dfrac{\partial M}{\partial t}+\sum_{i=1}^p\dfrac{\partial M}{\partial \hat{\theta}_i}\dot{\hat{\theta}}_i\right)\mathtt{e}+2\sum_{q=x,x_d}dM_{q}\dot{q}(t;\theta) \nonumber \\
=&(({d}/{dt})|_{\hat{\theta}}{M})\mathtt{e}-2\sum_{q=x,x_d}dM_{q}(\dot{q}(t;\hat{\theta})-\dot{q}(t;\theta))
\end{align}
where $\dot{q}(t;\vartheta)$ is $\dot{q}$ computed with $\theta=\vartheta$ in \eqref{adaptive_general} and \eqref{adaptive_general_d}, and $({d}/{dt})|_{\hat{\theta}}{M}$ is the time derivative of $M$ computed along \eqref{adaptive_general} and \eqref{adaptive_general_d} with $\theta = \hat{\theta}$. Thus, \eqref{linear_adaptive} of Assumption~\ref{multiplicative_asmp} gives $\dot{M}\mathtt{e}=(({d}/{dt})|_{\hat{\theta}}{M})\mathtt{e}-2(dM_x^{\top}Y+dM_{x_d}^{\top}Y_d)\tilde{\theta}$, resulting in $\dot{V}/2 \leq -\alpha_{\mathrm{NCM}} \mathtt{e}^{\top}M\mathtt{e}-\mathtt{e}^{\top}(dM_x^{\top}Y+dM_{x_d}^{\top}Y_d+M\tilde{Y})\tilde{\theta}+\tilde{\theta}^{\top}\Gamma\dot{\tilde{\theta}}+\mathtt{e}^{\top}Md$ as in the proof of Theorem~\ref{adaptive_affine_thm}, due to the relations \eqref{error_dynamics}, \eqref{adaptive_general_u}, and \eqref{deterministic_contraction_adaptive_tilde}. The adaptation law \eqref{adaptation_general} and the conditions \eqref{Merror_adaptive} and \eqref{Merror_adaptive_d} applied to this relation yield
\begin{align}
\dot{V}/2 \leq& -\alpha_{\mathrm{NCM}} \mathtt{e}^{\top}M\mathtt{e}+\bar{y}\epsilon_{\ell}\|\mathtt{e}\|\|\tilde{\theta}\|-\sigma\tilde{\theta}^{\top}\hat{\theta}+\mathtt{e}^{\top}Md \\
\leq&-(\alpha_{\mathrm{NCM}}/\overline{\omega})\|\mathtt{e}\|^2+\bar{y}\epsilon_{\ell}\|\mathtt{e}\|\|\tilde{\theta}\|-\sigma\|\tilde{\theta}\|^2+\bar{d}_{a}\sqrt{V}
\end{align}
for $\bar{d}_{a}$ in \eqref{adaptive_bound_1}, which implies $d\sqrt{V}/dt \leq -\alpha_{a}\sqrt{V}+\bar{d}_{a}$ by \eqref{condition_aNCM}. The rest follows from Theorem~\ref{adaptive_affine_thm}.
\end{proof}

The aNCM control of Theorem~\ref{adaptiveNCMthm} also has the following asymptotic stability property as in Corollary~\ref{Cor:adaptiveNCM}.
\begin{corollary}
\label{Cor:adaptiveNCM2}
The aNCM control \eqref{adaptive_general_u} with the adaptation \eqref{adaptation_general} guarantees $\lim_{t\to\infty}\|\mathtt{e}(t)\| = 0$ for $\mathtt{e}=x-x_d$ when $\epsilon_{\ell} = 0$, $d(x) = 0$, and $\sigma = 0$ in \eqref{adaptive_general}, \eqref{adaptation_general}, \eqref{Merror_adaptive}, and \eqref{Merror_adaptive_d}.
\end{corollary}
\begin{proof}
We have $\dot{V} \leq -2\alpha_{\mathrm{NCM}} \mathtt{e}^{\top}M\mathtt{e}$ in this case by the proof of Theorem~\ref{adaptiveNCMthm}. The rest follows from Corollary~\ref{Cor:adaptiveNCM}.
\end{proof}
\begin{remark}
As discussed in Remark~\ref{Remark:objective}, the steady-state error of \eqref{adaptive_bound_1} can also be used in \eqref{convex_opt_ancm} for optimal disturbance attenuation in an adaptive sense. The dependence on $u$ and $\dot{\hat{\theta}}$ in $(d/dt)|_{\hat{\theta}}M$ can be removed by using $\partial_{b_i(x)}M+\partial_{b_i(x_d)}M=0$ and using adaptation rate scaling introduced in~\cite{lopez2021universal}.
\end{remark}

\section{Practical Application of aNCM Control}
\label{applications}
This section derives one way to use the aNCM control with function approximators and proposes practical numerical algorithms to construct it using Theorems~\ref{adaptive_affine_thm}--\ref{adaptiveNCMthm}.
%%%%%%%%%%%%%%%%%%%%%%%%%%%%%%%%%%%%%%%
%%%%%%%%%%%%%%%%%%%%%%%%%%%%%%%%%%%%%%%
\subsection{Systems Modeled by Function Approximators}
Utilization of function approximators, neural networks, in particular, has gained great popularity in system identification due to their high representational power, and provably-stable techniques for using
these approximators in closed-loop have been derived in~\cite{SannerSlotine1992}. The aNCM adaptive control frameworks are applicable also in this context.

Suppose $f(q)$ and $B(q)=[b_1(q),\cdots,b_m(q)]$ of \eqref{sdc_dynamics} are modeled with the basis functions 
$\phi(q)=[\phi_1(q),\cdots,\phi_p(q)]^{\top}$ and 
$\varphi_i(q)=[\varphi_{i,1}(q),\cdots,\varphi_{i,q}(q)]^{\top}$, $i=1,\cdots,m$ for $q=x,x_d$:
\begin{align}
\label{basis_func_dyn}
\dot{q} =& f(q)+B(q)u = \mathsf{F}\phi(q)+\sum_{i=1}^m\mathsf{B}_i\varphi_i(q)u_i+d_M(q)
\end{align}
where $d_M(q)$ with $\sup_{x}\|d_M(q)\| = \bar{d}_M/2 < \infty$ is the modeling error, $\mathsf{F}\in\mathbb{R}^{n\times p}$, and $\mathsf{B}_i\in\mathbb{R}^{n\times q}$. Note that $\mathsf{F}$ and $\mathsf{B}_i$ are the ideal weights with small enough $\bar{d}_M$, but let us consider the case where we only have access to their estimates, $\hat{\mathsf{F}}$ and $\hat{\mathsf{B}}_i$ due to, \eg{}, insufficient amount of training data. Theorem~\ref{basis_function_prop} introduces the aNCM-based adaptation law to update $\hat{\mathsf{F}}$ and $\hat{\mathsf{B}}_i$ for exponential boundedness of the system trajectories.
\begin{theorem}
\label{basis_function_prop}
Let $\mathcal{M}(x,x_d,\hat{\mathsf{F}},\hat{\mathsf{B}})$ be the aNCM of Theorem~\ref{adaptiveNCMthm}, where $\hat{\mathsf{F}}$ and $\hat{\mathsf{B}}$ are the estimates of $\mathsf{F}$ and $\mathsf{B}$ in \eqref{basis_func_dyn}. Also, let $\mathsf{W}$ denote the weights $\mathsf{F}$ and $\mathsf{B}_i$, and define $\zeta$ and $\zeta_d$ as $\zeta=\phi(x)$ and $\zeta_d=\phi(x_d)$ for $\mathsf{W}=\mathsf{F}$, and $\zeta=\varphi_i(x)u_i$ and $\zeta_d=\varphi_i(x_d)u_{d,i}$ for $\mathsf{W}=\mathsf{B}_i$. Suppose that \eqref{basis_func_dyn} is controlled by $u$ of \eqref{adaptive_general_u} with the following adaptation law:
\begin{align}
\label{adaptation_F}
\dot{\hat{\mathsf{W}}} = \Gamma^{-1}\textbf{:} (d\mathcal{M}_x\mathtt{e}\zeta^{\top}+d\mathcal{M}_{x_d}\mathtt{e}\zeta_{d}^{\top}+\mathcal{M}\mathtt{e}\tilde{\zeta}^{\top}-\sigma\hat{\mathsf{W}})
\end{align}
where $\sigma \in [0,\infty)$, $\tilde{\zeta}=\zeta-\zeta_{d}$, $\textbf{:}$ is defined as $(A\textbf{:}B)_{ij}=\sum_{k,\ell}A_{ijk\ell}B_{\ell k}$, and $\Gamma$ represents the fourth order tensor given with $\underline{\gamma} \|S\|_F^2 \preceq S\textbf{:}\Gamma\textbf{:}S \preceq \overline{\gamma}\|S\|_F^2,~\forall S \in \mathbb{R}^p$ for $\underline{\gamma},\overline{\gamma} \in (0,\infty)$ and the Frobenius norm $\|S\|_F = S\textbf{:}S \leq \|S\|$. If $\exists \bar{\zeta}\in (0,\infty)$ \st{} $\|\zeta\|\leq\bar{\zeta}$, $\|\zeta_d\|\leq\bar{\zeta}$, and $\|\tilde{\zeta}\|\leq\bar{\zeta},~\forall x,x_d$, and if $\Gamma$ and $\sigma$ are selected to satisfy the following for $\epsilon_{\ell}$ of \eqref{Merror_adaptive} and \eqref{Merror_adaptive_d}:
\begin{align}
\label{condition_aNCMnet}
\begin{bmatrix}
-2\alpha_{\mathrm{NCM}}/\overline{\omega} & \bar{\zeta}\epsilon_{\ell}\bm{1}_{m+1}^{\top} \\
\bar{\zeta}\epsilon_{\ell}\bm{1}_{m+1} & -2\sigma I_{m+1}
\end{bmatrix} \preceq -2\alpha_a\begin{bmatrix}1/\underline{\omega} & 0 \\ 0 & (1/\underline{\gamma})I_{m+1}\end{bmatrix}
\end{align}
where $\alpha_a\in(0,\infty)$ and $\bm{1}_k = [1,\cdots,1]^{\top} \in \mathbb{R}^k$, then $\mathtt{e}=x-x_d$ of \eqref{basis_func_dyn} is exponentially bounded as in \eqref{adaptive_bound_1}. When $\epsilon_{\ell}=0$, $d_M=0$, and $\sigma = 0$ in \eqref{Merror_adaptive}, \eqref{Merror_adaptive_d}, \eqref{basis_func_dyn}, and \eqref{adaptation_F}, the system \eqref{basis_func_dyn} controlled by \eqref{adaptive_general_u} is asymptotically stable.
\end{theorem}
\begin{proof}
Let us define $V$ as $V = V_{\mathtt{e}}+\sum_{\mathsf{W}=\mathsf{F},\mathsf{B}_i}\tilde{\mathsf{W}}\textbf{:}\Gamma^{-1}\textbf{:}\tilde{\mathsf{W}}$ for $M$ in Theorem~\ref{adaptiveNCMthm}, where $V_{\mathtt{e}}=\mathtt{e}^{\top}M\mathtt{e}$ and $\tilde{\mathsf{W}}=\hat{\mathsf{W}}-\mathsf{W}$. Since $M$ is given by \eqref{deterministic_contraction_adaptive_tilde}, we have as in the proof of Theorem~\ref{adaptiveNCMthm} that $\dot{V}_{\mathtt{e}} \leq -2\alpha_{\mathrm{NCM}} \mathtt{e}^{\top}M\mathtt{e}-2\mathtt{e}^{\top}\sum_{\mathsf{W}=\mathsf{F},\mathsf{B}_i}(dM_x^{\top}\tilde{\mathsf{W}}\zeta+dM_{x_d}^{\top}\tilde{\mathsf{W}}\zeta_{d}+M\tilde{\mathsf{W}}\tilde{\zeta})$. Using the relation $a^{\top}Cb=C\textbf{:}(ab^{\top})$ for $a\in\mathbb{R}^n$, $b\in\mathbb{R}^p$, and $C\in\mathbb{R}^{n\times p}$, we get
\begin{align}
\dot{V}/2 \leq -\alpha_{\mathrm{NCM}} V_{\mathtt{e}}+\bar{\zeta}\epsilon_{\ell}\|\mathtt{e}\|\sum_{\mathsf{W}=\mathsf{F},\mathsf{B}_i}\|\tilde{\mathsf{W}}\|-\sigma\tilde{\mathsf{W}}:\hat{\mathsf{W}}+\mathtt{e}^{\top}M\tilde{d}_M \nonumber
\end{align}
for $\tilde{d}_M=d_M(x)-d_M(x_d)$ with $\|\tilde{d}_M\| \leq \bar{d}_M$. The rest follows from the proof of Theorem~\ref{adaptiveNCMthm} and Corollary~\ref{Cor:adaptiveNCM2} along with the condition \eqref{condition_aNCMnet} and $\tilde{\mathsf{W}}\textbf{:}\tilde{\mathsf{W}}=\|\tilde{\mathsf{W}}\|_F^2 \geq \|\tilde{\mathsf{W}}\|^2$, where $\|\cdot\|$ denotes the induced $2$-norm.
\end{proof}
\begin{remark}
\label{bregman:neuralnet}
For systems modeled by DNNs, we can utilize the same technique in Theorem~\ref{basis_function_prop} to adaptively update the weights of its last layer. Such over-parameterized systems can always be implicitly regularized using the Bregman divergence~\cite{boffi2020implicit} (see Sec.~\ref{Sec:bregman}).
\end{remark}
\if0
We can utilize the same technique for DNN modeled systems to adaptively update the weights of its last layer.
\begin{corollary}
\label{adaptiveNCMthm_NN}
Consider the following system, $\dot{x} = f
(x)+B(x)u = \mathsf{F}^*\varrho(x;\mathsf{F}_{1:L})+\sum_{i=1}^m\mathsf{B}_i^*\beta_{i}(x;\mathsf{B}_{1:L}^{(i)})u_i$, where $\varrho$ and $\beta_{i}$ are the nominal DNNs with the trained weights of the first $L$ layers, $\bar{\mathsf{F}}_{1:L}$ and $\bar{\mathsf{B}}_{1:L}^{(i)}$, and $\mathsf{F}^*$ and $\mathsf{B}_i^*$ are the true weights of their last layers. Then the aNCM control \eqref{adaptive_general_u} under the adaptation law \eqref{adaptation_F} with $\phi(x)$ and $\varphi_i(x)$ replaced by $\rho(x;\mathsf{F}_{1:L})$ and $\beta_i(x;\mathsf{B}_{1:L}^{(i)})$, respectively, guarantees asymptotic convergence of $e=x-x_d$ to $0$.
\end{corollary}
\begin{proof}
See the proof of Proposition~\ref{basis_function_prop}.
\end{proof}
\fi
%%%%%%%%%%%%%%%%%%%%%%%%%%%%%%%%%%
%%%%%%%%%%%%%%%%%%%%%%%%%%%%%%%%%%
\subsection{Additional Remarks in aNCM Implementation}
We propose several useful implementation techniques for the application of the provably stable and robust adaptive control frameworks in Theorems~\ref{adaptive_affine_thm}--\ref{basis_function_prop}.
%%%%%%%%%%%%%%%%%%%%%%%%%%%%%%%%%%
\subsubsection{Constraints as Loss Functions}
Instead of solving \eqref{convex_opt_ncm} and \eqref{convex_opt_ancm} for $\bar{W}$ to sample training data $\{(x,x_d,M)\}_{i=1}^N$, we could directly solve them for the DNN weights, regarding the constraints as loss functions for the network training as described in \cite{chuchu}. This still gives the exponential bound of \eqref{adaptive_bound_1}, as long as we can get sufficiently small $\epsilon_{\ell}$ of \eqref{Merror} which satisfies the conditions of Theorems~\ref{adaptive_affine_thm}~and~\ref{adaptiveNCMthm}.
%%%%%%%%%%%%%%%%%%%%%%%%%%%%%%%%%%
\subsubsection{Implicit Regularization}
\label{Sec:bregman}
Over-parametrized systems can be implicitly regularized using the Bregman divergence as mentioned in Remarks~\ref{bregman:remark}~and~\ref{bregman:neuralnet}. In particular, it enables satisfying $\theta^*=\text{arg}\min_{\vartheta\in A}\psi(\vartheta)$, where $\theta^*=\lim_{t\to\infty}\hat{\theta}$, $A$ is the set containing only parameters that interpolate the dynamics along the entire trajectory, and $\psi$ can be any strictly convex function~\cite{boffi2020implicit}. For example, we could use $\psi(\vartheta) = \|\vartheta\|_p$, leading to various regularization properties depending on the choice of $p$ (\eg{} sparsity when $p=1$).
%%%%%%%%%%%%%%%%%%%%%%%%%%%%%%%%%%
\subsubsection{aNCMs for Control Lyapunov Functions}
The aNCM can also be utilized as a Control Lyapunov Function (CLF)~\cite{lagros}. In particular, we consider a controller $u=u_d(x_d)+K^*(x,x_d)\mathtt{e}$ in \eqref{adaptive_general}, where $K^*(x,x_d)$ is given by
\begin{align}
\label{ancm_robust_control}
&(K^*,p^*) = \text{arg}\min_{K\in \mathbb{R}^{m\times n},p\in\mathbb{R}}\|K\mathtt{e}\|^2+p^2 \\
\label{stability_clf}
&\text{\st{} } \left.({d}/{dt})\right|_{\hat{\theta}}{\mathcal{M}}+2\sym{}(\mathcal{M}A+\mathcal{M}K) \leq -2\alpha \mathcal{M}+pI_n~~
\end{align}
which is convex when $(x,x_d)$ is given at time $t$.
\begin{proposition}
The convex optimization \eqref{ancm_robust_control} is always feasible due to the relaxation variable $p$. Theorem~\ref{adaptiveNCMthm} still holds if $2\alpha \mathcal{M} \succ \bar{p}^*I_n$ for $\bar{p}^*=\sup_{x,x_d}p^*$. Note that convex input constraints can be incorporated in the same way.
\end{proposition}
\begin{proof}
See~\cite{lagros}.
\end{proof}
%%%%%%%%%%%%%%%%%%%%%%%%%%%%%%%%%%
\subsubsection{Pseudocode for aNCM Construction}
We finally note that the aNCM can be constructed with the pseudocodes provided in~\cite{ncm,nscm}, using \eqref{convex_opt_ncm} and \eqref{convex_opt_ancm} of Theorems~\ref{modified_ncm_thm}~and~\ref{adaptiveNCMthm} as their sampling methodology in this case.

\section{Simulation}
\label{simulation}
\begin{figure} 
    \centering
    \includegraphics[width=20mm]{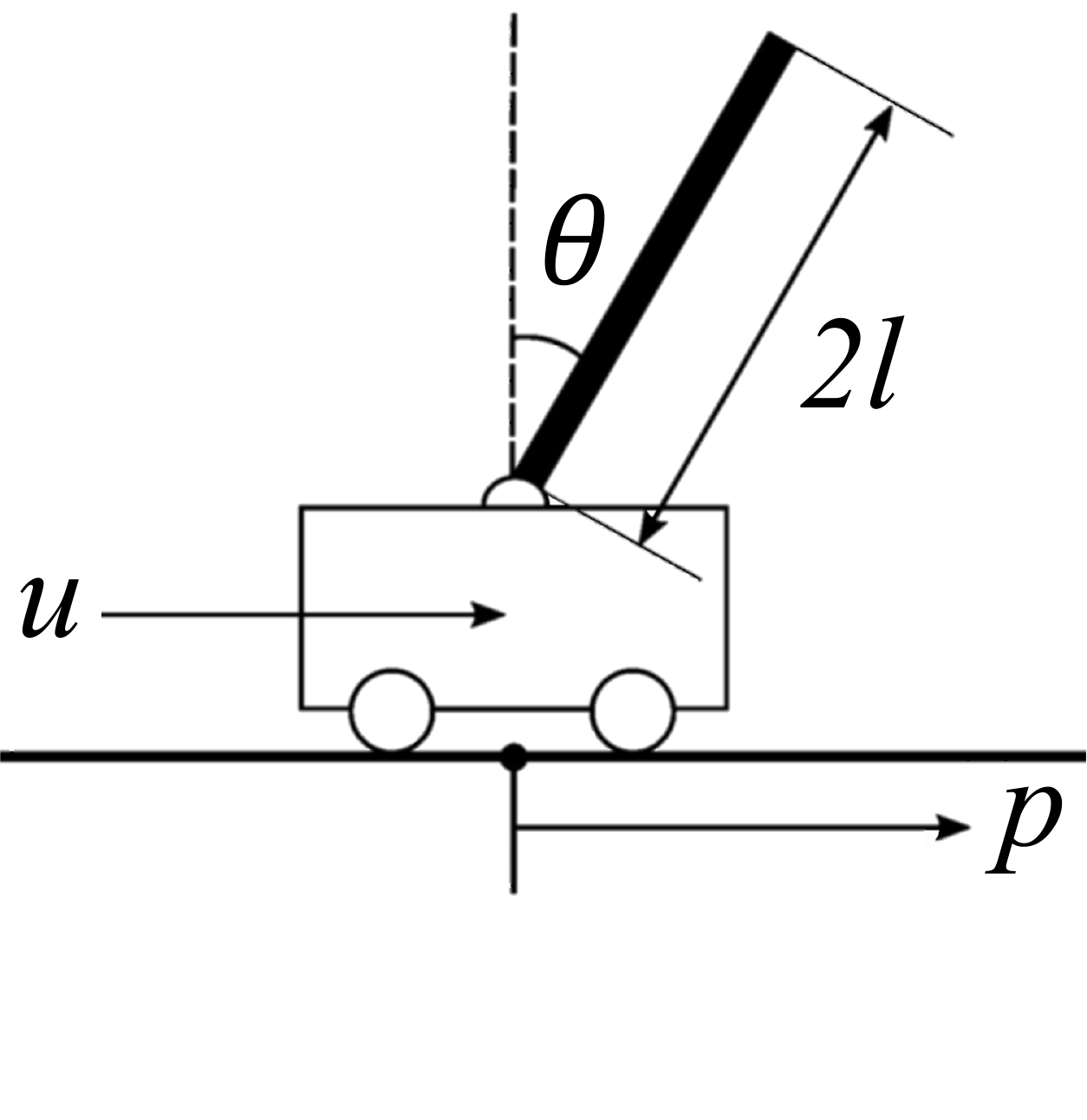}
    \vspace{-1.5em}
    \caption{Cart-pole balancing task.}
    \label{catpole_fig}
    \vspace{-1.5em}
\end{figure}

We demonstrate the aNCM framework in the cart-pole balancing problem~\cite{6313077} ({\color{caltechgreen}\underline{\href{https://github.com/astrohiro/ancm}{https://github.com/astrohiro/ancm}}}), where CVXPY~\cite{cvxpy} is used to solve convex optimization. The task is selected to drive the state $x=[p,\theta,\dot{p},\dot{\theta}]^{\top}$ in Fig.~\ref{catpole_fig} to $0$ controlling the under-actuated dynamics given as $(m_c+m)\ddot{p}+ml\cos\theta\ddot{\theta} = ml\dot{\theta}^2\sin\theta-\mu_c\dot{p}+u$, and $ml\cos\theta\ddot{p}+({4}/{3})ml^2\ddot{\theta} = mlg\sin\theta-\mu_p\dot{\theta}$,
%\begin{align}
%\label{cartpole_eq1}
%&\begin{bmatrix} m_c+m & ml\cos\theta \\
%ml\cos\theta & \frac{4}{3}ml^2
%\end{bmatrix}
%\begin{bmatrix} \ddot{p} \\ \ddot{\theta} \end{bmatrix}
%+
%\begin{bmatrix} 0 & -ml\dot{\theta}\sin\theta \\
%0 & 0
%\end{bmatrix}
%\begin{bmatrix} \dot{p} \\ \dot{\theta} \end{bmatrix} \\
%&+
%\begin{bmatrix} 0 \\ mlg\sin\theta \end{bmatrix}
%+
%\begin{bmatrix} \mu_c\dot{p} \\ \mu_p\dot{\theta} \end{bmatrix}
%=
%\begin{bmatrix} 1 \\ 0 \end{bmatrix}u
%\end{align}
%\begin{align}
%\label{cartpole_eq1}
%(m_c+m)\ddot{p}+ml\cos\theta\ddot{\theta} =& ml\dot{\theta}^2\sin\theta-\mu_c\dot{p}+u \\
%\label{cartpole_eq2}
%ml\cos\theta\ddot{p}+\frac{4}{3}ml^2\ddot{\theta} =& mlg\sin\theta-\mu_p\dot{\theta}
%\end{align}
%\begin{align}
%\label{cartpole_eq1}
%\ddot{\theta} =& \frac{g\sin\theta+\cos\theta\left(\frac{-u-ml \dot{\theta}^2\sin\theta+\mu_c\dot{p}}{m_c+m}\right)-\frac{\mu_p\dot{\theta}}{ml}}{l\left(\frac{4}{3}-\frac{m\cos^2\theta}{m_c+m}\right)} \\
%\label{cartpole_eq2}
%\ddot{p} =& \frac{u+ml (\dot{\theta}^2\sin\theta-\ddot{\theta}\cos\theta)-\mu_c\dot{p}}{m_c+m}
%\end{align}
where $g = 9.8$, $m_c = 1.0$, $m = 0.1$, $\mu_c = 0.5$, $\mu_p = 0.002$, and $l = 0.5$. Note that the systems in this section are perturbed by the disturbance $d(x)$ with $\sup_x\|d(x)\|=0.15$.
%%%%%%%%%%%%%%%%%%%%%%%%%%%%%%%%%%%%%%%
%%%%%%%%%%%%%%%%%%%%%%%%%%%%%%%%%%%%%%%
%%%%%%%%%%%%%%%%%%%%%%%%%%%%%%%%%%%%%%%
\subsubsection{Neural Network Training}
We use a DNN of $\mathcal{M}$ with $3$ layers and $100$ neurons. The DNN is trained using stochastic gradient descent with training data sampled by \eqref{convex_opt_ncm} and \eqref{convex_opt_ancm} of Theorems~\ref{modified_ncm_thm} and \ref{adaptiveNCMthm} ($10000$ training samples), and the loss function is defined as in \cite{ncm}.
%%%%%%%%%%%%%%%%%%%%%%%%%%%%%%%%%%%%%%%
%%%%%%%%%%%%%%%%%%%%%%%%%%%%%%%%%%%%%%%
%%%%%%%%%%%%%%%%%%%%%%%%%%%%%%%%%%%%%%%
\subsubsection{Cart-Pole Balancing with Unknown Drags}
Let us first consider the case where $\mu_c$ and $\mu_p$ are unknown, which satisfies Assumption~\ref{multiplicative_asmp} to apply the aNCM in Theorem~\ref{adaptiveNCMthm}. Although the matching condition in Theorem~\ref{adaptive_affine_thm} does not hold, \eqref{affine_adaptive_u} is also implemented using the pseudo-inverse of $B(x)$ in \eqref{adaptive_affine}. The adaptive robot trajectory control~\cite[pp. 403]{Slotine:1228283} is not applicable as the dynamics is under-actuated, and thus we use it for partial feedback linearization as in (68) of~\cite{mypaperTAC}. We compare their performance with the iterative LQR (iLQR)~\cite{ilqr} and robust NCM in Theorem~\ref{modified_ncm_thm} without any adaptation. The initial conditions are selected as $x(0)=[0.83,-0.32,0.39,0.45]^{\top}$, $\hat{\mu}_c(0)=4$, and $\hat{\mu}_p(0)=0.0016$.
\begin{figure}
    \centering
    \includegraphics[width=80mm]{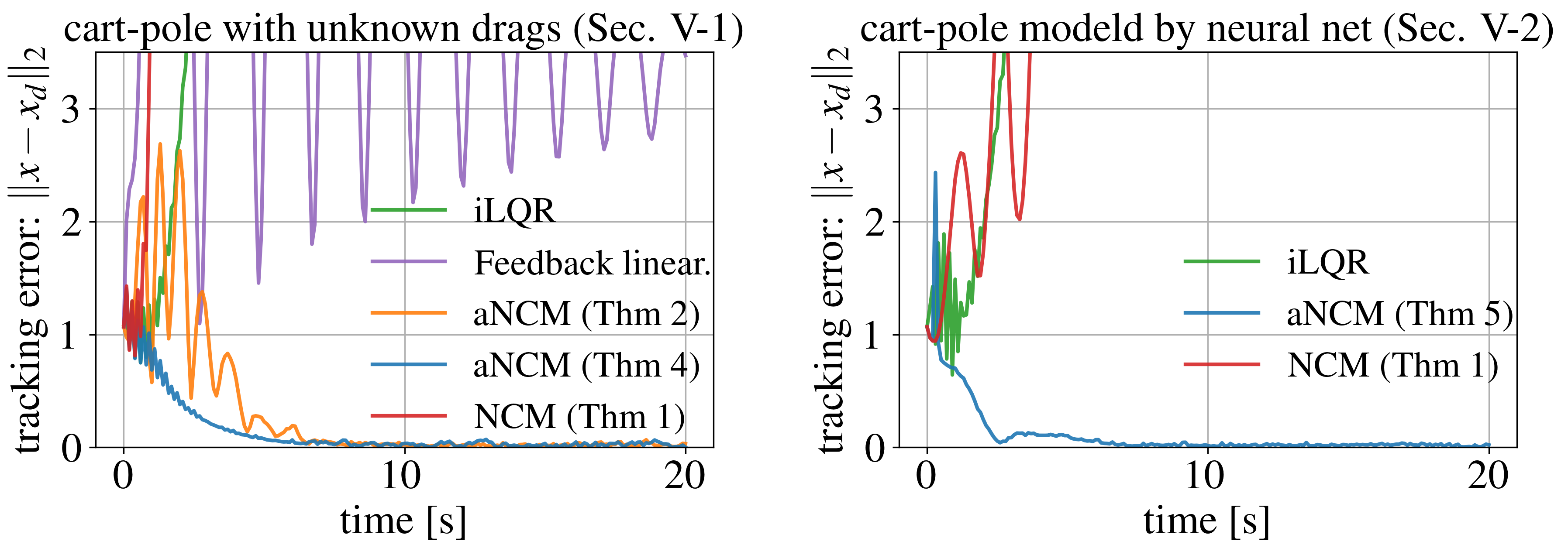}
    \vspace{-1.2em}
    \caption{Simulation results for cart-pole balancing task with unknown drags (LHS) and unknown dynamical system (RHS).}
    \label{ancm_sim_results}
    \vspace{-1.5em}
\end{figure}

As can be seen from Fig.~\ref{ancm_sim_results}, the aNCM control law of Theorems~\ref{adaptive_affine_thm}~and~\ref{adaptiveNCMthm} achieve stabilization, while the other three baselines in \cite[pp. 403]{Slotine:1228283}, \cite{ncm}, and \cite{ilqr} fail to balance the pole. Also, the aNCM of Theorem~\ref{adaptiveNCMthm} has a better transient behavior than that of Theorem~\ref{adaptive_affine_thm} as the matched uncertainty condition does not hold in this case.
%%%%%%%%%%%%%%%%%%%%%%%%%%%%%%%%%%%%%%%
%%%%%%%%%%%%%%%%%%%%%%%%%%%%%%%%%%%%%%%
%%%%%%%%%%%%%%%%%%%%%%%%%%%%%%%%%%%%%%%
\subsubsection{Cart-Pole Balancing with Unknown Dynamical System}
\label{sim:NNcartpole}
We next consider the case where the structure of the cart-pole dynamics is unknown and modeled by a DNN with $3$ layers and $5$ neurons, assuming we have $10000$ training samples generated by the true dynamics. Its modeling error is set to a relatively large value, $0.5$, so we can see how the proposed adaptive control achieves stabilization even for such poorly modeled dynamics. The performance of the aNCM control in Theorem~\ref{basis_function_prop} is compared with that of the iLQR~\cite{ilqr} and baseline robust NCM control in Theorem~\ref{modified_ncm_thm} constructed for the nominal DNN dynamical system model.

As shown in the right-hand side of Fig.~\ref{ancm_sim_results}, the proposed aNCM control indeed achieves stabilization even though the underlying dynamical system is unknown, while the trajectories of the iLQR and robust NCM computed for the nominal DNN dynamical system diverge.

\section{Conclusion}
\label{conclusion}
This work presents the method of aNCM, which uses a DNN-based differential Lyapunov function to provide formal stability and robustness guarantees for nonlinear adaptive control, even in the presence of parametric uncertainties, external disturbances, and aNCM learning errors. It is applicable to a wide range of systems including those modeled by neural networks and demonstrated to outperform existing robust and adaptive control in Sec.~\ref{simulation}. Using it with~\cite{nscm,lagros} would also enable adaptive motion planning under stochastic perturbation. By using a DNN, the aNCM framework presents a promising direction for obtaining formal stability guarantees of adaptive controllers without resorting to real-time numerical computation of a Lyapunov function. 
%\subsubsection*{Acknowledgments}
%This work was supported in part by the Jet Propulsion Laboratory, California Institute of Technology. 

\bibliographystyle{IEEEtran}
\bibliography{root}

\end{document}